# Bangla Text Classification using Transformers

Tanvirul Alam
*BJIT Limited*
Dhaka, Bangladesh
tanvirul.alam@bjitgroup.com

Akib Khan
*BJIT Limited*
Dhaka, Bangladesh
akib.khan@bjitgroup.com

Firoj Alam
*Qatar Computing Research Institute, HBKU*
Doha, Qatar
fialam@hbku.edu.qa

*Abstract*—Text classification has been one of the earliest problems in NLP. Over time the scope of application areas has broadened, and the difficulty of dealing with new areas (e.g., noisy social media content) has increased. The problem-solving strategy switched from classical machine learning to deep learning algorithms. One of the recent deep neural network architecture is the Transformer. Models designed with this type of network and its variants recently showed their success in many downstream natural language processing tasks, especially for resource-rich languages, e.g., English. However, these models have not been explored fully for Bangla text classification tasks. In this work, we fine-tune multilingual transformer models for Bangla text classification tasks in different domains, including *sentiment analysis*, *emotion detection*, *news categorization*, and *authorship attribution*. We obtain the state of the art results on six benchmark datasets, improving upon the previous results by 5-29% accuracy across different tasks.

## I. INTRODUCTION

Text classification is a classic topic in natural language processing (NLP) with many real-world applications. It refers to the task of classifying textual units such as sentences, queries, paragraphs, and documents into predefined labels or tags. Some applications of text classification include sentiment analysis, news categorization, user intent classification, content moderation, etc.

Some of the common sources of data for text classification are web pages, social media, online news portal, emails, online shops, user reviews, and questions and answers from customer services. Even though there is an abundance of textual data, preparing such data for the classification task is not only challenging but also time-consuming due to its unstructured and noisy nature.

Earlier works on text classification were based on classical machine learning algorithms. This required manually selecting features like the bag of words or n-grams, which were then used as inputs to classification algorithms such as Naive Bayes (NB), Support Vector Machine (SVM), Hidden Markov Models (HMM), random forests [1]–[3], etc. However, as large scale datasets [4] have become available in recent years, the focus has been shifted to use deep learning algorithms. As deep learning models can learn representation from the data itself, it reduces the need for feature engineering and makes the models transferable across different tasks.

Distributed word representations learned using neural networks have been widely used as they are capable of learning rich semantics using large unlabeled data [5], [6]. These word embeddings are then used to classify the texts using different neural networks like Multi-layer Perceptrons (MLP), Convolutional Neural Networks (CNN), and Recurrent Neural Networks (RNN) [7]–[10].

More recently, transformers [11] based pre-trained language models have been successfully used for learning language representations by utilizing a large amount of unlabeled data. Some of these models include OpenAI GPT [12], BERT [13], RoBERTa [14]. These models have proven to be immensely successful when fine-tuned on different downstream tasks, including text classification [15], question answering [16], natural language inference [17], etc. However, these models are usually trained on large monolingual English corpora or on multilingual corpora that can include over a hundred languages.

Recent work has shown that fine-tuning from multi-lingual models can achieve comparable performance to monolingual models [18] for low resource languages. Motivated by this, and the fact that, there has been no prior work on Bangla text classification based on this approach, we explore the efficacy of different multilingual models for Bangla text classification tasks. We experiment on six publicly available datasets with diverse topics including *sentiment analysis*, *emotion detection*, *authorship attribution* and *news categorization*. We show the effectiveness of the proposed approach by obtaining state of the art results on them.

We organize the rest of the paper as follows. In Section II, we discuss works related to text classification in Bangla language. We briefly discuss the datasets used in our study in Section III. Our proposed approach and training details are described in Section IV. We compare our results with previous work and draw meaningful insight into our approach in Section V. Finally, in Section VI, we conclude the paper.

## II. RELATED WORKS

Compared to English the work on Bangla text classification is limited in spite of being one of the most spoken and culturally rich languages with nearly 265 million native speakers. The main reason is the scarcity of labeled data for training the machine learning models. In this section, we highlight the works that are relevant to our work.

*a) Sentiment/Emotion Classification:* For sentiment and emotion classification the current state of the art for Bangla

includes resource development and addressing the model development challenges. The earlier work includes rule-based and classical machine learning algorithms. In [19], the authors propose a computational technique of generating an equivalent SentiWordNet (Bangla) from publicly available English sentiment lexicons and English-Bangla bilingual dictionary with very few easily adaptable noise reduction techniques. The classical algorithms used in different studies include Bernoulli Naive Bayes (BNB), Decision Tree, SVM, Maximum Entropy (ME), and Multinomial Naive Bayes (MNB) [20]–[22]. In [23], the authors developed a polarity detection system on textual movie reviews in Bangla by using two popular machine learning algorithms such as NB and SVM and provided comparative results. In another study, authors used NB with rules for detecting sentiment in Bengali Facebook Status [23]. In [24], the authors developed a dataset using semi-supervised approaches and designed models using SVM and Maximum Entropy [24].

The work related to the use of deep learning algorithms for sentiment analysis include [25]–[29]. In [27], the authors used Long Short Term Memory (LSTM) and CNN with embedding layer for both sentiment and emotion identification from youtube comments. The study in [28], provides a comparative analysis using both classical – SVM, and deep learning algorithms – LSTM and CNN, for sentiment classification in Bangla news comments. The study in [29] integrated word embeddings into a Multichannel Convolutional-LSTM (MConv-LSTM) network for predicting different types of hate speech, document classification, and sentiment analysis for Bangla. Due to the availability of romanized Bangla texts in social media, the studies in [25], [26] use LSTM to design and evaluate the model for sentiment analysis. In [30], authors used CNN sentiment classification in Bangla comments. The studies in [31] and [32] analyze user sentiment on Cricket comments from online news forums.

*b) Authorship identification:* Authorship attribution is another interesting research problem in which the task is to identify original authors from the text. The research work in this area is comparatively low. In [33], the authors developed a dataset and experiment with character level embedding for Authorship Attribution.

*c) New Classification:* A large dataset of Bangla articles from different news portals, which contains around 3,76,226 articles were provided [34]. The study conducted experiments using Logistic Regression, Neural Network, NB, Random Forest, and Adaboost by utilizing textual features such as word2vec, TF-IDF(3000 word vector), and TF-IDF(300 word vector). In [35], the authors extracted tf-idf features and performed Bangla content classification using Random Forest, SVM with linear and radial basis kernel, K-Nearest Neighbor, Gaussian Naive Bayes, and Logistic Regression. They have created a large Bangla content dataset and made it publicly available.

TABLE I: Statistics of the datasets used in the experiments. C: number of classes, L: average text length (words), Train/Dev/Test: number of samples in the respective splits (if provided officially)

| Dataset | C | Train | Dev | Test | L |
|---|---|---|---|---|---|
| YouTube Sentiment-3 [27] | 3 | 8910 | - | - | 11 |
| YouTube Sentiment-5 [27] | 5 | 3886 | - | - | 11 |
| YouTube Emotion [27] | 5 | 2890 | - | - | 10 |
| News Comment Sentiment [28] | 5 | 13802 | - | - | 20 |
| Authorship Attribution [33] | 14 | 14047 | - | 3511 | 750 |
| News Classification [36] | 6 | 11284 | 1411 | 1411 | 231 |

## III. DATASETS

We experiment with multiple publicly available datasets. These datasets are collected from four sources and include diverse topics such as sentiment classification, emotion detection, authorship attribution, news categorization. Statistics of different datasets used are shown in Table I.

### A. YouTube comment datasets

Three datasets are collected from YouTube user comments in [27]. Two of these are for sentiment analysis and one for emotion detection. One sentiment analysis dataset consists of 5 classes: strongly positive, positive, neutral, negative, and strongly negative and the other one has 3 classes: positive, neutral, and negative. The emotion detection dataset has 5 types of emotion: anger/disgust, joy, sadness, fear/surprise, and none. One interesting aspect of these datasets is that they contain texts in Bangla, English, and romanized Bangla.

### B. News comment sentiment dataset

This sentiment analysis dataset was developed in [28] from Bangla news portals. It has five categories of sentiments: slightly positive, definitely positive, neutral, slightly negative, and definitely negative. For training, slightly positive and definitely positives comments were considered as positive, while slightly negative and definitely negative comments were considered as negative, and neutral comments were dropped in the paper.

### C. Authorship Attribution dataset

This dataset in [33] contains writings of 14 different authors from online Bangla e-library (e.g., novels, story, series, etc.). Each document in the dataset has fixed length of 750 words.

### D. News Classification dataset

This dataset was prepared for the news classification task in [36]. It contains 6 different classes of interest. The authors provide train, validation, and test split for this dataset.

## IV. EXPERIMENTS

### A. Models and Architecture

We fine-tune multi-lingual transformer models that are trained on large corpus. We have not used monolingual Bangla model as there is no such model publicly available to the best of our knowledge. A general architecture of our model is shown in Figure 1.

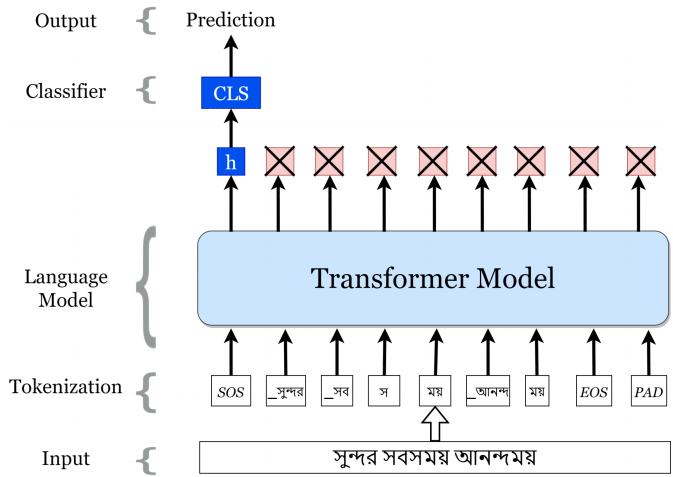

Fig. 1: Model Architecture

We use model specific tokenizers to split the input text into a list of tokens. As these models use byte-pair encoding [37], single word may be split into multiple tokens. For example, here the input sentence সুন্দর সবসময় আনন্দময় *(Beauty is joy forever)* is split into 6 tokens when using the XLM-RoBERTa-large tokenizer.

*1) Pretrained Language Models:* We use transformer language models in our experiments, which are available publicly in HuggingFace's transformer library [38]. To fine-tune the transformer model for the classification task, we insert a special *start of sequence (SOS)* token at the start and a special *end of sequence (EOS)* token at the end.[1] As we use fixed-size input to the models, we add padding at the end if necessary or remove extra tokens from the end if the number of tokens exceed the fixed sequence length. Padding tokens are masked out so they are not present during training. These tokens form the input for the transformer models where they are passed through multiple self-attention layer, and form the final hidden embedding corresponding to each input token. For the classification task, we only consider the first token corresponding to the start of the sequence token. This is then used to produce final output probability distribution over the number of categories.

We experiment with three models from two model classes: multilingual BERT and XLM-RoBERTa.

*a) Multilingual BERT:* BERT [13] is trained to learn distributed representation from unlabeled texts by jointly conditioning on left and right contexts of a token. It uses the encoder part of the Transformer architecture introduced in [11]. Two objective functions are used during the language model pretraining step. The first one is masked language model (MLM) that randomly masks[2] some fraction of the input tokens and the objective is to predict the vocabulary

[1]For BERT the *SOS* and *EOS* tokens are respectively [CLS] and [SEP]. For XLM-RoBERTa these are <s> and </s>.
[2]A special token [MASK] is used for this

ID of the original token in that position. The bidirectional nature ensures that the model can effectively make use of both past and future tokens for this. The second objective is the next sentence prediction (NSP) task. This is a binary classification task for predicting whether two sentences are subsequent in the original text. Positive sentences are created by taking consecutive sentences from the text and negative sentences are created by taking sentences from two different documents.

The multilingual variant of BERT (mBERT) is trained on more than 100 languages with the largest Wikipedia corpus. Since different languages have a different amount of Wikipedia entries, data is sampled using an exponentially smoothed weighting (with a factor 0.7). This ensures that high resource languages like English are under-sampled compared to low resource languages. Word counts are also sampled in a similar manner so that low resource languages have sufficient words in the vocabulary.

*b) XLM-RoBERTa:* RoBERTa [14] improves upon BERT by training on larger datasets, using larger vocabulary, and training on longer sequences with larger batches. NSP task is removed and only MLM loss is used for pretraining.

XLM-RoBERTa [18] is the multilingual variant of RoBERTa trained with a multilingual MLM. It is trained on one hundred languages, with more than two terabytes of filtered Common Crawl data. XLM-RoBERTa showed impressive performance in several multilingual NLP tasks and can perform comparably to monolingual language models.

*2) Classification Head:* We add a task-specific classification head for fine-tuning the model for specific tasks. The hidden representation obtained from the start of sequence token can be treated as the sentence embedding and used for classification [13], [14]. This gives us a $H$ dimensional embedding vector. For BERT, we add a linear layer with neurons equal to the number of classes for the task. For RoBERTa, a hidden layer is used with $H$ neurons with $tanh$ non-linearity and is followed by the final classification layer.

### B. Training Procedure

We trained the models using cross entropy loss criterion and Adam optimization algorithm [39]. All models were trained for 10 epochs with learning rate 1e-5. We use 32 samples in each mini batch, except when this does not fit in memory (e.g., XLM-RoBERTa large model trained on authorship attribution dataset). In such cases we use maximum batch size that fits within GPU memory.[3] We used fixed sequence length during training and apply padding or truncate when necessary. The sequence length consists of 30, 100, 300, and 300 tokens for YouTube comments, News Comments, Authorship, News datasets, respectively. All model parameters are fine-tuned during training i.e., no layer is kept frozen. The model with the best validation set performance is evaluated on the test dataset.

[3]The models were trained on a single 16 GB NVIDIA Tesla P100 GPU

TABLE II: Result on YouTube sentiment (3 class) dataset

|  | Model | Accuracy | F1 Score |
|---|---|---|---|
| Others | LSTM [27] | **66.0** | **63.5** |
|  | CNN [27] | 60.9 | 60.5 |
|  | NB [27] | 60.8 | 59.5 |
|  | SVM [27] | 59.2 | 58.9 |
| Ours | BERT-base | 71.7 | 71.5 |
|  | XLM-RoBERTa-base | 74.4 | 74.2 |
|  | XLM-RoBERTa-large | **76.0** | **75.9** |

TABLE III: Result on YouTube sentiment (5 class) dataset

|  | Model | Accuracy | F1 Score |
|---|---|---|---|
| Others | LSTM [27] | **54.2** | **53.2** |
|  | CNN [27] | 52.1 | 52.1 |
|  | NB [27] | 46.9 | 48.0 |
|  | SVM [27] | 44.9 | 46.5 |
| Ours | BERT-base | 53.5 | 52.8 |
|  | XLM-RoBERTa-base | 57.4 | 56.7 |
|  | XLM-RoBERTa-large | **60.4** | **59.8** |

TABLE IV: Result on YouTube emotion detection dataset

|  | Model | Accuracy | F1 Score |
|---|---|---|---|
| Others | LSTM [27] | **59.2** | **52.9** |
|  | CNN [27] | 54.0 | 53.5 |
|  | NB [27] | 52.5 | 52.5 |
|  | SVM [27] | 49.3 | 49.8 |
| Ours | BERT-base | 60.4 | 59.1 |
|  | XLM-RoBERTa-base | 69.8 | 66.6 |
|  | XLM-RoBERTa-large | **72.7** | **70.6** |

TABLE V: Result on news comment sentiment dataset

|  | Model | Accuracy | F1 Score |
|---|---|---|---|
| Others | SVM [28] | 61.34 | 63.97 |
|  | LSTM [28] | **74.74** | **79.29** |
|  | CNN [28] | 60.49 | 66.24 |
| Ours | BERT-base | 74.68 | 72.32 |
|  | XLM-RoBERTa-base | 76.61 | 74.69 |
|  | XLM-RoBERTa-large | **78.64** | **78.41** |

TABLE VI: Result on authorship attribution dataset

|  | Model | Accuracy | F1 Score |
|---|---|---|---|
| Others | Char-CNN [33] | 69.0 | - |
|  | W2V(CBOW) [33] | 71.8 | - |
|  | fastText(CBOW) [33] | 40.3 | - |
|  | W2V(Skip) [33] | 78.6 | - |
|  | fastText(Skip) [33] | **81.2** | - |
| Ours | BERT-base | 82.6 | 82.8 |
|  | XLM-RoBERTa-base | 87.2 | 87.2 |
|  | XLM-RoBERTa-large | **93.8** | **93.8** |

TABLE VII: Result on news classification dataset

|  | Model | Accuracy |  |
|---|---|---|---|
| Others | FT-W [36] | 62.79 | - |
|  | FT-WC [36] | 64.78 | - |
|  | INLP [36] | **72.50** | - |
| Ours | BERT-base | 91.28 | 91.28 |
|  | XLM-RoBERTa-base | 92.70 | 92.84 |
|  | XLM-RoBERTa-large | **93.41** | **93.40** |

## V. RESULTS AND DISCUSSIONS

### A. Results on different Datasets

In this section, we describe the evaluation procedures used for specific dataset and compare results obtained using transformer models to those previously reported. We evaluate the models using accuracy and weighted F1 score.

*1) YouTube comment datasets:* The results on YouTube sentiment (3 and 5 class) and emotion detection datasets are reported in Table II, III and IV. The baseline SVM and Naive Bayes (NB) models in [27] were trained using tf-idf features with n-gram tokens, while CNN and LSTM models were trained using word embedding learned using word2vec [5]. We follow similar procedure for reporting result as in their work. Specifically, we set aside 10% of each dataset for testing. The rest was further divided into 80% for training and 20% for validation. We repeated the experiments 10 times and report average result for each model.

It is evident that transformer models consistently yield better performance on these datasets compared to the benchmark models, only exception being BERT on the 5-class sentiment dataset where it performs slightly worse. Most significant improvement is observed on the emotion detection dataset where XLM-RoBERTa large achieves 22.8% relative increase in accuracy compared to the LSTM model.

*2) News Comment Sentiment:* We omitted the neutral class and only consider binary sentiment classification for this dataset as was done in [28]. We split the dataset into train, validation and test splits having 80%, 10% and 10% samples respectively. The accuracy and F1 score are reported in Table V. The authors in [28] reported results for SVM, CNN and LSTM models. Transformer models perform better than SVM and CNN models and comparable to the LSTM model in this dataset.

*3) Authorship Attribution:* We balanced the dataset prior to training, taking minimum number of samples (469) per class similar to [33]. We used the train and test splits provided in [33] for the experiments and split 20% of the training data further into validation set. We limit sequence length to 300 on this dataset even though each sample in the dataset has 750 words. This was done to meet GPU memory constraint.

The authors in [33] reported results for character and word level CNN models trained with fastText and word2vec embedding and best results were obtained with skip-gram variants of fastText embedding. All three transformer models perform better than this model with XLM-RoBERTa-large model significantly outperforming with a 15.5% relative increase in accuracy as reported in Table VI.

*4) News classification:* We used the train, validation and test splits provided in [36] for news classification. Results obtained on this dataset is reported in Table VII. The authors in [36] used fastText embedding trained on different corpora to train KNN classifier. This means further improvement could be gained by training CNN or LSTM model from the embedding. Regardless, we notice significant improvement using transformer models in this dataset. All three variants of our models achieve greater than 25% relative increase in accuracy compared to the best benchmark model [36].

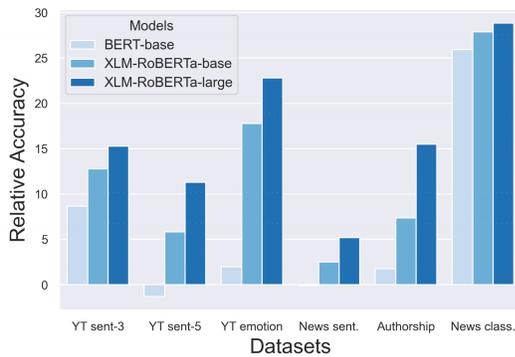

Fig. 2: Relative accuracy compared to the best published results on different datasets

*B. Discussions*

We can conclude from the results that transformer based models are better suited for Bangla text classification tasks compared to classical machine learning based approaches that use manually engineered features or CNN/LSTM models trained on distributed word representation. We represent the relative accuracy of the three models compared to the best published prior results on each dataset in Figure 2. There is a clear trend across the datasets. XLM-RoBERTa-base performs better than BERT and XLM-RoBERTa-large performs better than XLM-RoBERTa-base model. Both XLM-RoBERTa models improve upon previously reported results across all datasets. BERT performs slightly worse than the previous best result on YouTube sentiment-5 class and News sentiment datasets.

There are various factors contributing to the performance gain obtained by XLM-RoBERTa models. They are trained on a larger corpus compared to BERT and has larger vocabulary size. As a result fewer unknown tokens are introduced after tokenization. For example, considering the example sentence from before, we obtain following tokens after using XLM-RoBERTa model tokenizer

'_সুন্দর', '_সব', 'স', 'ময়', '_আনন্দ', 'ময়'

However, using BERT tokenizer we get the following tokens
'স', '##ুন', '##দ', '##র', '[UNK]', '[UNK]'

Last two words are replaced with the unknown tokens, even though they are not rare word in Bangla. The first word is also split into several tokens. XLM-RoBERTa performs better in this regard due to availability of larger vocabulary for Bangla in the model. As XLM-RoBERTa-large is a model with trained with more layers and parameters, it also has better generalization ability than XLM-RoBERTa-base model. This results in better performance in the downstream tasks.

One advantage of subword embedding is that it can perform better in noisy user generated text like those found in online

(a) Word importance for sentence 'কেন জানি না! গানটা এতো ভালো লাগে। সবসময় শুনতে ভালো লাগে।' *(I don't know why! I like the song so much. Always good to* hear.) Detected as positive with 0.9972 probability.

(b) Word importance for sentence 'শেষের সিনটা টা খুব খারাপ লেগেছিল *(The last scene was very bad!)* Detected as negative with 0.9627 probability.

(c) Word importance for sentence 'লাইফে এমন গান খুব কম শুনেছি,হাজার বার শুনেও খারাপ লাগেনি' *(I have rarely heard such songs in my life, I have not felt bad even after listening to them a thousand times.)* Detected as positive with 0.9807 probability.

Fig. 3: Example sentences highlighting the importance of words that network learns.

media. As they tend to contain misspelling and shortened words, subword embedding (also character embedding) can still capture some meaningful information from them. People also often comment in both Bangla and English in those platforms, as is present in the YouTube datasets. Since we are using multilingual models for fine-tuning which are also trained on large amount of English data, the model is expected to perform better in such cases. This can explain why we got significantly more improvement from transformer models in the YouTube datasets compared to the Bangla news sentiment dataset, which does not contain any English text.

Using transformer models for fine-tuning also reduces the need for feature engineering and data preprocessing. In our experiments, we did not use any separate text preprocessing steps (e.g., stop words and punctuation removal) other than using the model specific tokenizers.

In this experiment, we have explored multilingual language models as they are more readily available, especially for a low resource language like Bangla. If we fine-tune from a monolingual Bangla transformer model pretrained on a large corpus, we believe further improvements can be made on Bangla only texts like in authorship attribution dataset.

For a better interpretation of the model we explore and show word importance with three example sentences. This is obtained for XLM-RoBERTa large model trained on YouTube sentiment-3 dataset. These representations are produced using Captum [40]. In Figure 3a, the model puts a high emphasis on the positive word ভালো (good). First sentence কেন জানি না! (I don't know why!) can be considered to be slightly negative and is reflected in the word importance. In Figure 3b, the negative word খারাপ (bad) is properly highlighted to arrive at the overall negative sentiment. Interestingly খুব (very) is highlighted as opposite i.e., positive, meaning the model was not able to combine two words and emphasize the phrase খুব খারাপ (very bad) as negative. Finally, the sentence in Figure 3c is more challenging. However the model is able to identify the overall positive tone despite the presence of some negative words like কম, খারাপ (rarely, bad).

## VI. CONCLUSIONS

We have explored different transformer models for a variety of Bangla text classification tasks. Our work shows that fine-tuning from transformer models can yield better performance compared to traditional methods that make use of hand-crafted features, and also deep learning models like CNN and LSTM trained on distributed word representation. We obtained a state of the art result on six benchmark datasets from different domains. We hope this will encourage researchers to make use of these models for other tasks in the Bangla language.

## REFERENCES

[1] L. M. Manevitz and M. Yousef, "One-class SVMs for document classification," *Journal of machine Learning research*, vol. 2, no. Dec, pp. 139–154, 2001.

[2] B. Pang, L. Lee, and S. Vaithyanathan, "Thumbs up? Sentiment classification using machine learning techniques," in *Proc. of EMNLP*, 2002.

[3] G. Forman, "An extensive empirical study of feature selection metrics for text classification," *Journal of machine learning research*, vol. 3, no. Mar, pp. 1289–1305, 2003.

[4] X. Zhang, J. Zhao, and Y. LeCun, "Character-level convolutional networks for text classification," in *Advances in neural information processing systems*, 2015, pp. 649–657.

[5] T. Mikolov, K. Chen, G. Corrado, and J. Dean, "Efficient estimation of word representations in vector space," in *1st International Conference on Learning Representations, ICLR 2013, Scottsdale, Arizona, USA, May 2-4, 2013, Workshop Track Proceedings*, 2013.

[6] J. Pennington, R. Socher, and C. D. Manning, "Glove: Global vectors for word representation," in *EMNLP*, 2014.

[7] Y. Kim, "Convolutional neural networks for sentence classification," *arXiv preprint arXiv:1408.5882*, 2014.

[8] K. S. Tai, R. Socher, and C. D. Manning, "Improved semantic representations from tree-structured long short-term memory networks," *arXiv preprint arXiv:1503.00075*, 2015.

[9] S. Wan, Y. Lan, J. Guo, J. Xu, L. Pang, and X. Cheng, "A deep architecture for semantic matching with multiple positional sentence representations," *arXiv preprint arXiv:1511.08277*, 2015.

[10] A. Joulin, E. Grave, P. Bojanowski, and T. Mikolov, "Bag of tricks for efficient text classification," *arXiv preprint arXiv:1607.01759*, 2016.

[11] A. Vaswani, N. Shazeer, N. Parmar, J. Uszkoreit, L. Jones, A. N. Gomez, L. u. Kaiser, and I. Polosukhin, "Attention is all you need," in *Advances in Neural Information Processing Systems 30*. Curran Associates, Inc., 2017, pp. 5998–6008. [Online]. Available: http://papers.nips.cc/paper/7181-attention-is-all-you-need.pdf

[12] A. Radford, K. Narasimhan, T. Salimans, and I. Sutskever, "Improving language understanding by generative pre-training," *URL https://s3-us-west-2. amazonaws. com/openai-assets/research-covers/language-unsupervised/language_ understanding_paper. pdf*, 2018.

[13] J. Devlin, M.-W. Chang, K. Lee, and K. Toutanova, "BERT: Pre-training of deep bidirectional transformers for language understanding," *arXiv preprint arXiv:1810.04805*, 2018.

[14] Y. Liu, M. Ott, N. Goyal, J. Du, M. Joshi, D. Chen, O. Levy, M. Lewis, L. Zettlemoyer, and V. Stoyanov, "RoBERTa: A robustly optimized BERT pretraining approach," *CoRR*, vol. abs/1907.11692, 2019.

[15] C. Sun, X. Qiu, Y. Xu, and X. Huang, "How to fine-tune BERT for text classification?" in *China National Conference on Chinese Computational Linguistics*. Springer, 2019, pp. 194–206.

[16] S. Garg, T. Vu, and A. Moschitti, "TANDA: Transfer and adapt pre-trained transformer models for answer sentence selection," *arXiv preprint arXiv:1911.04118*, 2019.

[17] Z. Zhang, Y. Wu, H. Zhao, Z. Li, S. Zhang, X. Zhou, and X. Zhou, "Semantics-aware BERT for language understanding," *arXiv preprint arXiv:1909.02209*, 2019.

[18] A. Conneau, K. Khandelwal, N. Goyal, V. Chaudhary, G. Wenzek, F. Guzmán, E. Grave, M. Ott, L. Zettlemoyer, and V. Stoyanov, "Unsupervised cross-lingual representation learning at scale," *arXiv preprint arXiv:1911.02116*, 2019.

[19] A. Das and S. Bandyopadhyay, "SentiWordNet for Bangla," *Knowledge Sharing Event-4: Task*, vol. 2, pp. 1–8, 2010.

[20] A. Rahman and M. S. Hossen, "Sentiment analysis on movie review data using machine learning approach," in *2019 International Conference on Bangla Speech and Language Processing (ICBSLP)*, 2019, pp. 1–4.

[21] N. Banik and M. H. H. Rahman, "Evaluation of naïve bayes and support vector machines on Bangla textual movie reviews," in *Proc. of ICBSLP*. IEEE, 2018, pp. 1–6.

[22] R. R. Chowdhury, M. S. Hossain, S. Hossain, and K. Andersson, "Analyzing sentiment of movie reviews in Bangla by applying machine learning techniques," in *Proc. of (ICBSLP)*. IEEE, 2019, pp. 1–6.

[23] M. S. Islam, M. A. Islam, M. A. Hossain, and J. J. Dey, "Supervised approach of sentimentality extraction from Bengali facebook status," in *Proc. of ICCIT*. IEEE, 2016, pp. 383–387.

[24] S. Chowdhury and W. Chzowdhury, "Performing sentiment analysis in Bangla microblog posts," in *2014 International Conference on Informatics, Electronics Vision (ICIEV)*, 2014, pp. 1–6.

[25] A. Hassan, M. R. Amin, A. K. Al Azad, and N. Mohammed, "Sentiment analysis on Bangla and romanized Bangla text using deep recurrent models," in *2016 International Workshop on Computational Intelligence (IWCI)*. IEEE, 2016, pp. 51–56.

[26] A. A. Sharfuddin, M. N. Tihami, and M. S. Islam, "A deep recurrent neural network with BiLSTM model for sentiment classification," in *Proc. of ICBSLP*. IEEE, 2018, pp. 1–4.

[27] N. I. Tripto and M. E. Ali, "Detecting multilabel sentiment and emotions from Bangla youtube comments," in *2018 International Conference on Bangla Speech and Language Processing (ICBSLP)*. IEEE, 2018, pp. 1–6.

[28] M. A.-U.-Z. Ashik, S. Shovon, and S. Haque, "Data set for sentiment analysis on bengali news comments and its baseline evaluation," in *Proc. of ICBSLP*. IEEE, 2019, pp. 1–5.

[29] M. R. Karim, B. R. Chakravarthi, J. P. McCrae, and M. Cochez, "Classification benchmarks for under-resourced Bengali language based on multichannel convolutional-lstm network," *CoRR*, vol. abs / 2004.07807, 2020.

[30] M. H. Alam, M.-M. Rahoman, and M. A. K. Azad, "Sentiment analysis for Bangla sentences using convolutional neural network," in *Proc. of ICCIT*. IEEE, 2017, pp. 1–6.

[31] M. Rahman, E. Kumar Dey *et al.*, "Datasets for aspect-based sentiment analysis in Bangla and its baseline evaluation," *Data*, vol. 3, no. 2, p. 15, 2018.

[32] S. A. Mahtab, N. Islam, and M. M. Rahaman, "Sentiment analysis on Bangladesh cricket with support vector machine," in *2018 International Conference on Bangla Speech and Language Processing (ICBSLP)*. IEEE, 2018, pp. 1–4.

[33] A. Khatun, A. Rahman, M. S. Islam *et al.*, "Authorship attribution in bangla literature using character-level CNN," in *Proc. of ICCIT*. IEEE, 2019, pp. 1–5.

[34] M. Tanvir Alam and M. Mofijul Islam, "BARD: Bangla article classification using a new comprehensive dataset," in *Proc. of ICBSLP*, 2018, pp. 1–5.

[35] S. Al Mostakim, F. Ehsan, S. Mahdiea Hasan, S. Islam, and S. Shatabda, "Bangla content categorization using text based supervised learning methods," in *2018 International Conference on Bangla Speech and Language Processing (ICBSLP)*, 2018, pp. 1–6.

[36] A. Kunchukuttan, D. Kakwani, S. Golla, G. N.C., A. Bhattacharyya, M. M. Khapra, and P. Kumar, "Ai4bharat-indicnlp corpus: Monolingual corpora and word embeddings for indic languages," *arXiv preprint arXiv:2005.00085*, 2020.

[37] R. Sennrich, B. Haddow, and A. Birch, "Neural machine translation of rare words with subword units," *arXiv preprint arXiv:1508.07909*, 2015.

[38] T. Wolf, L. Debut, V. Sanh, J. Chaumond, C. Delangue, A. Moi, P. Cistac, T. Rault, R. Louf, M. Funtowicz, J. Davison, S. Shleifer, P. von Platen, C. Ma, Y. Jernite, J. Plu, C. Xu, T. L. Scao, S. Gugger, M. Drame, Q. Lhoest, and A. M. Rush, "Huggingface's transformers: State-of-the-art natural language processing," *ArXiv*, vol. abs/1910.03771, 2019.

[39] D. P. Kingma and J. Ba, "Adam: A Method for Stochastic Optimization," in *3rd International Conference on Learning Representations*, 2015.

[40] N. Kokhlikyan, V. Miglani, M. Martin, E. Wang, B. Alsallakh, J. Reynolds, A. Melnikov, N. Kliushkina, C. Araya, S. Yan, and O. Reblitz-Richardson, "Captum: A unified and generic model interpretability library for pytorch," 2020.